\documentclass[conference]{IEEEtran}
\IEEEoverridecommandlockouts

\usepackage{cite}
\usepackage{amsmath,amssymb,amsfonts,bm}
\usepackage{graphicx}
\graphicspath{{Figures/}}
\usepackage{booktabs}
\usepackage{array}
\usepackage{xcolor}
\usepackage{url}

\newcommand{\ours}{\textsc{OURS-LITE}}
\newcommand{\fullobs}{\textsc{FULL OBS}}
\newcommand{\localnogate}{\textsc{LOCAL-NO-GATE}}
\newcommand{\euclidean}{\textsc{EUCLIDEAN}}
\newcommand{\ctde}{\textsc{CTDE MAPPO}}

\begin{document}

\title{Less is More: Robust Zero-Communication 3D Pursuit-Evasion via Representational Parsimony}

\author{
    \IEEEauthorblockN{
        Jialin Ying, 
        Zhihao Li, 
        Zicheng Dong, 
        Guohua Wu and 
        Yihuan Liao\IEEEauthorrefmark{1}
    }
    \\ \vspace{1ex}
    \IEEEauthorblockA{
        Department of Automation, Central South University, Changsha, China
    }
    \thanks{*Corresponding author: Yihuan Liao (email: 225189@csu.edu.cn)}
}
\maketitle
\sloppy

\begin{abstract}
Asymmetric 3D pursuit--evasion in cluttered voxel environments is difficult under communication latency, partial observability, and nonholonomic maneuver limits. While many MARL methods rely on richer inter-agent coupling or centralized signals, these dependencies can become fragility sources when communication is delayed or noisy. Building on an inherited path-guided decentralized pursuit scaffold \cite{pgf_mappo_2025}, we study a robustness-oriented question: can representational parsimony improve communication-free coordination?

We instantiate this principle with (i) a parsimonious actor observation interface that removes team-coupled channels (83-D $\rightarrow$ 50-D), and (ii) Contribution-Gated Credit Assignment (CGCA), a locality-aware credit structure for communication-denied cooperation. In Stage-5 evaluation (4 pursuers vs. 1 evader), our configuration reaches $0.753\pm0.091$ success and $0.223\pm0.066$ collision, outperforming the 83-D \fullobs\ counterpart ($0.721\pm0.071$, $0.253\pm0.089$). It further shows graceful degradation under speed/yaw/noise/delay stress tests and resilient zero-shot transfer on urban-canyon maps (about 61\% success at density 0.24). These results support a practical design principle: in communication-constrained multi-robot pursuit, representational sparsity may be more robust than richer inter-agent coupling.
\end{abstract}


\section{Introduction}
3D pursuit--evasion in cluttered voxel spaces is difficult because communication latency, partial observability, and nonholonomic constraints interact with highly non-convex geometry. In the 4-vs-1 setting considered here, pursuers are maneuver-limited ($v_x\!\le\!8$ m/s, $|\dot\psi|\!\le\!0.8$ rad/s nominally) while the evader reaches 9 m/s in Stage 5, so stale estimates and delayed corrections can destabilize cooperative interception.

A common MARL strategy is to increase inter-agent coupling via richer teammate-conditioned observations, explicit communication, or centralized critics. Although helpful in nominal simulation, these couplings may propagate stale peer beliefs under delay/noise. We therefore ask: \emph{does decentralized coordination in complex 3D pursuit necessarily require richer cross-agent coupling?}

Reference \cite{pgf_mappo_2025} introduced the path-guided collaborative search-and-capture scaffold (3D A* guidance + directional frontier allocation). Rather than revisiting that scaffold design, this paper studies a different question: which observation and credit structures remain robust under communication denial, delay, and sensing noise in cluttered 3D pursuit--evasion? We treat the inherited learner as a parameter-shared decentralized PPO/IPPO-style actor--critic (as implemented in \cite{pgf_mappo_2025}), distinct from centralized-critic MAPPO variants.

Our core insight is a robustness-oriented design principle: \emph{representational parsimony can improve communication-free coordination}. Our study highlights two counter-intuitive effects in this regime: (i) more team-coupled information can reduce rather than improve robustness; and (ii) communication-free cooperation can remain strong when sparse local observations are paired with locality-aware credit assignment. Our contributions are:
\begin{itemize}
    \item \textbf{Representational parsimony as a design principle for decentralized MARL in communication-constrained 3D pursuit.} We identify that reducing explicit cross-agent observation coupling can improve robustness under delay and sensing noise.
    \item \textbf{Contribution-Gated Credit Assignment (CGCA).} We introduce a locality-aware credit structure that sustains cooperative capture quality without explicit communication channels.
    \item \textbf{Comprehensive robustness/generalization evidence.} We provide benchmark evaluation plus speed/yaw/noise/delay stress suites and zero-shot transfer on procedurally generated urban-canyon maps.
\end{itemize}
The macro-topological planning scaffold itself is retained from \cite{pgf_mappo_2025}; our novelty lies in robustness-oriented information and credit design on that inherited decentralized foundation.

\section{Related Work}
\subsubsection{\textit{A. 3D Pursuit--Evasion}}
Classical pursuit guidance is grounded in analytic controllers such as APF-style reactive potential shaping \cite{khatib1986apf} and proportional navigation (PN) \cite{zarchan2012pn}. These methods are lightweight and interpretable, but they are fundamentally local and therefore vulnerable to topological deadlocks in voxelized 3D clutter: the controller can remain trapped in obstacle-induced local minima or oscillatory turn--overshoot cycles. Topology-aware planning (e.g., A*) alleviates some deadlocks by exposing homotopy-level structure \cite{hart1968astar}, yet standalone planning remains brittle under delayed actuation and model mismatch. Recent DRL pipelines improve adaptability, but naive end-to-end formulations frequently face combinatorial explosion in high-dimensional partially observable state spaces \cite{mappo2022}. The path-guided decentralized scaffold of \cite{pgf_mappo_2025} (named PGF-MAPPO in that paper) therefore motivates a pragmatic compromise: a planner encodes global geometry, while the policy specializes in local, uncertainty-aware residual control.

\subsubsection{\textit{B. Communication-Constrained MARL}}
Explicit communication in MARL has been extensively studied through differentiable messaging, attention routing, and graph interaction modules, including CommNet, attentional communication, and targeted multi-agent communication \cite{commnet2016,atoc2018,tarmac2019,dgn2020}. In simulation, these channels can improve coordination efficiency; however, high-frequency aerial pursuit is precisely the regime where communication quality is non-stationary. Delay, packet dropout, and asynchronous clocks convert message passing into stale belief propagation, which can trigger cross-agent error cascade rather than cooperation gains. This mismatch becomes especially acute when control horizons are short and collision margins are thin. Our zero-communication actor design is thus not an austerity choice; it is a robustness prior that explicitly severs fragile coupling pathways and forces policy learning to rely on local geometry plus shared topological guidance.

\subsubsection{\textit{C. Credit Assignment in Cooperative MARL}}
Credit assignment remains central in cooperative MARL. Counterfactual baselines such as COMA and value-factorization methods such as VDN/QMIX reduce multi-agent non-stationarity by introducing centralized structure during training \cite{valueDecomp2018,vdn2018,qmix2018}. These methods are powerful when global state is reliable, but the dependence on centralized signals can become a practical bottleneck in realistic distributed robotics, where synchronized state aggregation is costly or unavailable. In contrast, CGCA uses only local distance and closing-kinematics signals to shape cooperative incentives. This lightweight locality makes the mechanism naturally compatible with strict decentralized execution and avoids adding another high-dimensional centralized channel on top of already difficult 3D pursuit dynamics.

\section{Problem Formulation and Environment Setup}
\subsection{Stage-5 environment}
The world is discretized into a $52\times52\times18$ voxel grid (voxel size: 6.0 m) over $311.0\times311.0\times110.5$ m. Stage-5 settings include 60 m visibility range, 8 m capture radius, and a 70\% visibility gate. The traversable occupancy ratio is 0.127, and the largest free connected component covers 0.951 of free voxels, indicating cluttered but globally navigable topology.

\begin{table}[htbp]
\caption{Stage-5 environment and asymmetry constants.}
\label{tab:env}
\centering
\footnotesize
\begin{tabular}{@{}p{0.38\linewidth}p{0.57\linewidth}@{}}
\toprule
\textbf{Item} & \textbf{Value} \\
\midrule
World/grid & $311.0\times311.0\times110.5$ m, $52\times52\times18$ voxels \\
Team setup & 4 pursuers vs. 1 evader \\
Capture radius (Stage 5) & 8.0 m \\
Visibility range (Stage 5) & 60.0 m \\
Pursuer bounds & $|v_x|\le8$, $|v_y|\le4$, $|v_z|\le3$ m/s \\
Pursuer yaw cap & $|\dot\psi|\le0.8$ rad/s (swept down to 0.2) \\
Evader speed schedule & up to 9.0 m/s \\
Altitude occupancy anisotropy & 15.9\% / 12.1\% / 10.2\% \\
\bottomrule
\end{tabular}
\end{table}

\subsection{Asymmetric game model}
Let pursuer $i\in\{1,2,3,4\}$ have position $\bm p_t^i\in\mathbb R^3$, yaw $\psi_t^i$, and body-frame control
\begin{equation}
\bm u_t^i=[v_{x,t}^i,v_{y,t}^i,v_{z,t}^i,\dot\psi_t^i]^\top.
\end{equation}
Dynamics are
\begin{equation}
\bm p_{t+1}^i=\bm p_t^i+\Delta t\,\mathbf R_z(\psi_t^i)\bm v_t^i,\;
\psi_{t+1}^i=\psi_t^i+\Delta t\,\dot\psi_t^i,
\end{equation}
with constraints
\begin{equation}
|v_{x,t}^i|\le8,\;|v_{y,t}^i|\le4,\;|v_{z,t}^i|\le3,\;|\dot\psi_t^i|\le\bar\omega.
\end{equation}
Capture is defined by
\begin{equation}
\min_i\|\bm p_t^i-\bm p_t^e\|_2\le R_c,\quad R_c=8\,\text{m}.
\end{equation}
The cooperative objective is $J(\theta)=\mathbb E_{\pi_\theta}[\sum_t\gamma^t r_t]$ with dense collision/safety penalties and capture-quality incentives.

\begin{figure*}[t]
\centering
\includegraphics[width=\linewidth]{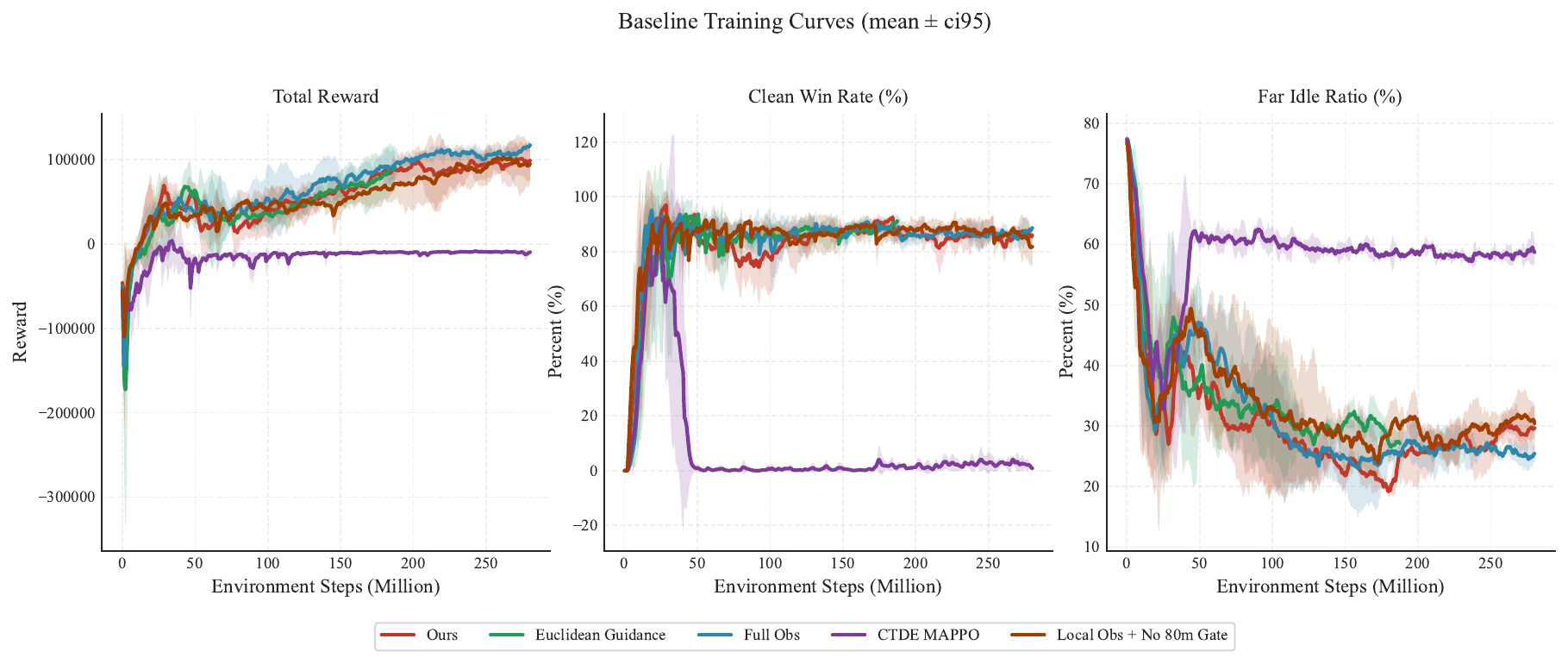}
\caption{Training reward curves under Stage 5. \ctde\ shows marked instability after the visibility-gated transition, whereas \ours\ converges to the highest stable return regime.}
\label{fig:train}
\end{figure*}

\section{Methodology}
\subsection{Inherited path-guided decentralized scaffold}
Following \cite{pgf_mappo_2025}, each agent receives a topology-aware 3D A* guidance vector instead of raw path polylines. The planner provides macro free-space structure, while the recurrent policy handles local reactive control. We retain this macro--micro decoupled pursuit scaffold as inherited infrastructure rather than a contribution of the present paper.

For method-boundary clarity, we follow the inherited implementation in \cite{pgf_mappo_2025} as a parameter-shared decentralized PPO/IPPO-style actor--critic, and treat centralized-critic MAPPO variants separately.

\subsection{Observation-space formalization: representational parsimony via 83-D to 50-D masking}
Let $\tilde{\bm o}_i^t\in\mathbb R^{83}$ denote the full actor-side observation for pursuer $i$ at time $t$:
\begin{equation}
\tilde{\bm o}_i^t=
\big[
\bm\ell_i^t,\Delta\bm p_{ie}^t,\Delta\bm v_{ie}^t,\bm v_i^t,\bm a_i^t,\bm\omega_i^t,
\phi(\mathcal N_i^t),\bm g_i^t,\bm e_i,m_t,\delta_t,\bm s_i^t,\bm c_i^t
\big],
\end{equation}
where $\phi(\mathcal N_i^t)\in\mathbb R^{24}$ encodes top-$k$ teammate states, $\bm s_i^t\in\mathbb R^{7}$ is the tactical-slot block, and $\bm c_i^t\in\mathbb R^{2}$ contains encirclement topology cues. To enforce representational parsimony under communication denial, we apply a binary masking operator $\mathbf M\in\{0,1\}^{50\times83}$:
\begin{equation}
\bm o_i^t=\mathbf M\tilde{\bm o}_i^t
=\big[\tilde{\bm o}_{i,1:41}^t,\tilde{\bm o}_{i,66:74}^t\big]\in\mathbb R^{50}.
\end{equation}
Equivalently, the operator explicitly nulls team-coupled channels,
\begin{equation}
\mathbf M\phi(\mathcal N_i^t)=\bm0,\qquad
\mathbf M\bm s_i^t=\bm0,\qquad
\mathbf M\bm c_i^t=\bm0,
\end{equation}
thereby removing 24-D teammate states and 9-D slot/encirclement descriptors. This operation is not introduced as generic feature pruning; it explicitly removes team-coupled channels that are most vulnerable to stale peer estimates, reducing sensitivity to delay/noise while keeping the action space and optimization backbone unchanged.

\begin{figure}[htbp]
\centering
\includegraphics[width=\linewidth]{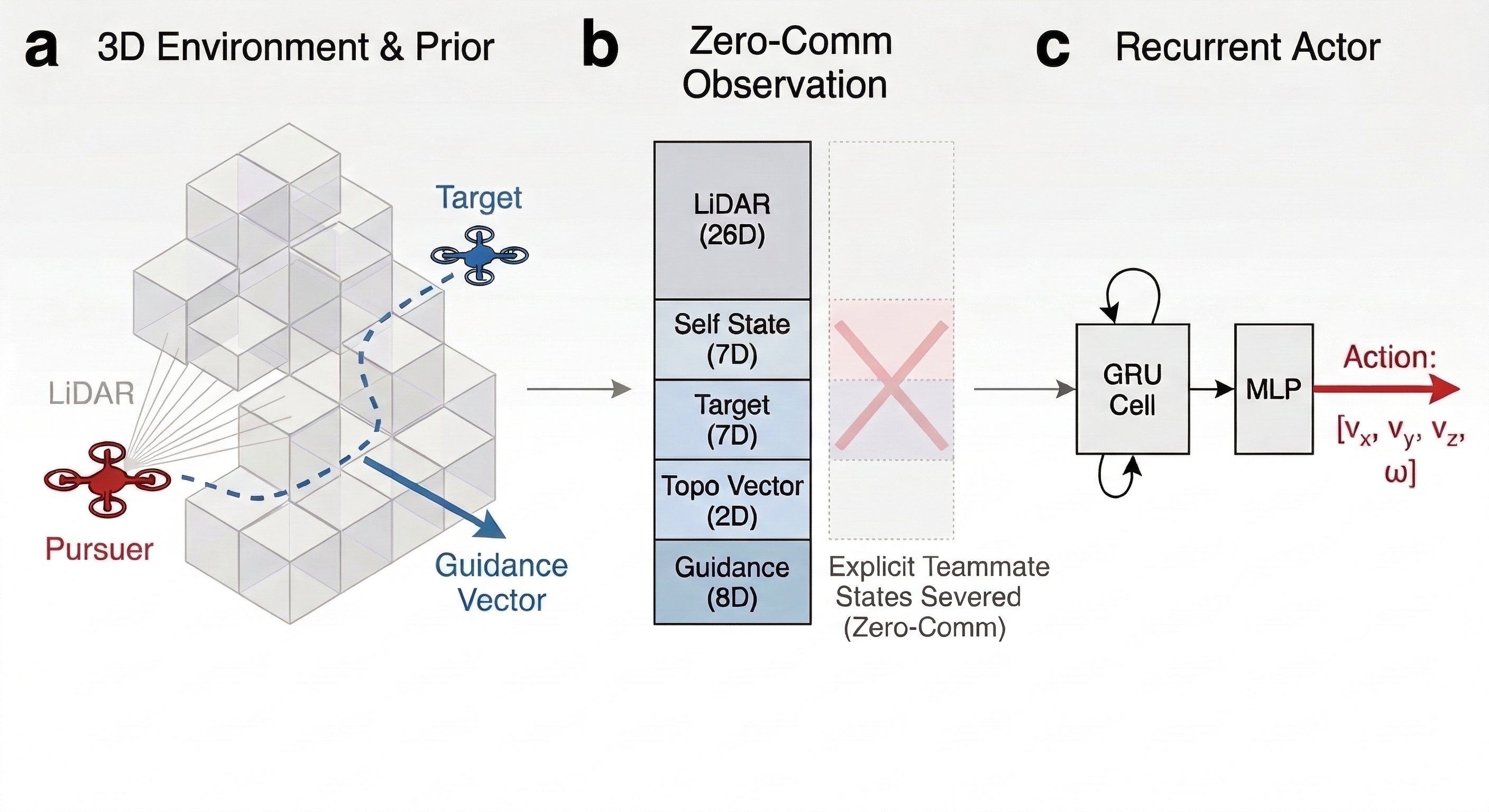}
\caption{Inherited path-guided decentralized pipeline \cite{pgf_mappo_2025}. Our change is at the actor interface: 83-D full observation is masked to 50-D by removing teammate/slot/encirclement channels for zero-communication robustness.}
\label{fig:arch}
\end{figure}

\begin{table}[htbp]
\caption{Observation-profile reduction for representational parsimony.}
\label{tab:obs}
\centering
\footnotesize
\begin{tabular}{lcc}
\toprule
\textbf{Block} & \textbf{83-D} & \textbf{50-D} \\
\midrule
LiDAR + target + self + IMU & 41 & 41 \\
Teammate top-$k$ & 24 & 0 \\
Guidance + ID + mode + delay & 9 & 9 \\
Slot + encirclement & 9 & 0 \\
\midrule
\textbf{Total} & 83 & \textbf{50} \\
\bottomrule
\end{tabular}
\end{table}

\subsection{Reward composition and Contribution-Gated Credit Assignment (CGCA)}
CGCA is designed as a locality-aware credit structure for communication-free cooperation, rather than a generic reward bonus. It rewards geometrically meaningful local participation in interception and suppresses free-rider equilibria when explicit messaging is unavailable.

The cooperative objective for parameters $\theta$ is
\begin{equation}
J(\theta)=\mathbb E_{\pi_\theta}\Bigg[\sum_{t=0}^{T}\gamma^t\frac{1}{N_t}\sum_{i=1}^{N_t}r_i^t\Bigg],
\end{equation}
with per-agent reward decomposition
\begin{align}
r_i^t ={}& \lambda_{\mathrm{dir}}\,g_{\mathrm{dir}}(d_i^t)\,\Delta d_i^t
+\lambda_{\mathrm{cap}}\,\rho^t\,\hat c_i^t\,\mathbf 1_{\mathrm{cap}}^t
+\lambda_{\mathrm{qual}}\,q_i^t \nonumber\\
&-\lambda_{\mathrm{col}}\,\mathbf 1_{\mathrm{col},i}^t
-\lambda_{\mathrm{imp}}\,\kappa_i^t
-\lambda_{\mathrm{lazy}}\,\mathbf 1_{\mathrm{lazy},i}^t,
\end{align}
where $\Delta d_i^t=d_i^{t-1}-d_i^t$, $d_i^t=\|\bm p_i^t-\bm p_e^t\|_2$, and $\kappa_i^t$ denotes collision-impact intensity.

To stabilize cooperation under zero communication, CGCA introduces locality-aware directional gating:
\begin{equation}
g_{\mathrm{dir}}(d)=
\begin{cases}
1,& d\le40\,\mathrm{m},\\
\frac{80-d}{40},&40<d\le80\,\mathrm{m},\\
0,& d>80\,\mathrm{m}.
\end{cases}
\end{equation}
Capture-share credit is hard-gated outside 60 m and weighted by closing behavior. Define closing speed $v_{i,\mathrm{clo}}^t=-\dot d_i^t$. The raw contribution is
\begin{equation}
\tilde c_i^t=\mathbf 1[d_i^t\le60]\Big(\alpha_p e^{-d_i^t/d_0}+\alpha_v[v_{i,\mathrm{clo}}^t]_++\alpha_r\mathbf 1[d_i^t\le R_c]\Big),
\end{equation}
and the normalized share is
\begin{equation}
\hat c_i^t=\frac{\tilde c_i^t}{\sum_{j=1}^{N_t}\tilde c_j^t+\epsilon}.
\end{equation}
To suppress free-rider behavior, a participation ratio multiplies global capture reward:
\begin{equation}
\eta_i^t=\mathbf 1[d_i^t\le60]\,\mathbf 1[v_{i,\mathrm{clo}}^t>0.5],\qquad
\rho^t=\min\!\left(1,\frac{\sum_i\eta_i^t}{0.5N_t}\right).
\end{equation}
Thus, if fewer than half of alive pursuers actively close the target, collective capture bonus is proportionally downscaled.

Collision suppression is explicitly encoded as
\begin{equation}
\mathbf 1_{\mathrm{col},i}^t=
\mathbf 1_{\mathrm{obs-hit},i}^t+
\mathbf 1_{\mathrm{team-hit},i}^t+
\mathbf 1_{\mathrm{shield},i}^t,
\end{equation}
which penalizes obstacle impacts, inter-agent contacts, and safety-shield triggers in a unified term.

The 40/60/80 m thresholds are chosen to align with the 60 m sensing extremum in Stage 5 and the practical maneuver envelope of the platform: 40 m acts as high-confidence local-interaction core, 60 m matches reliable observability scale, and 80 m suppresses non-contributive far-field reward leakage.

\begin{figure}[htbp]
\centering
\includegraphics[width=\linewidth]{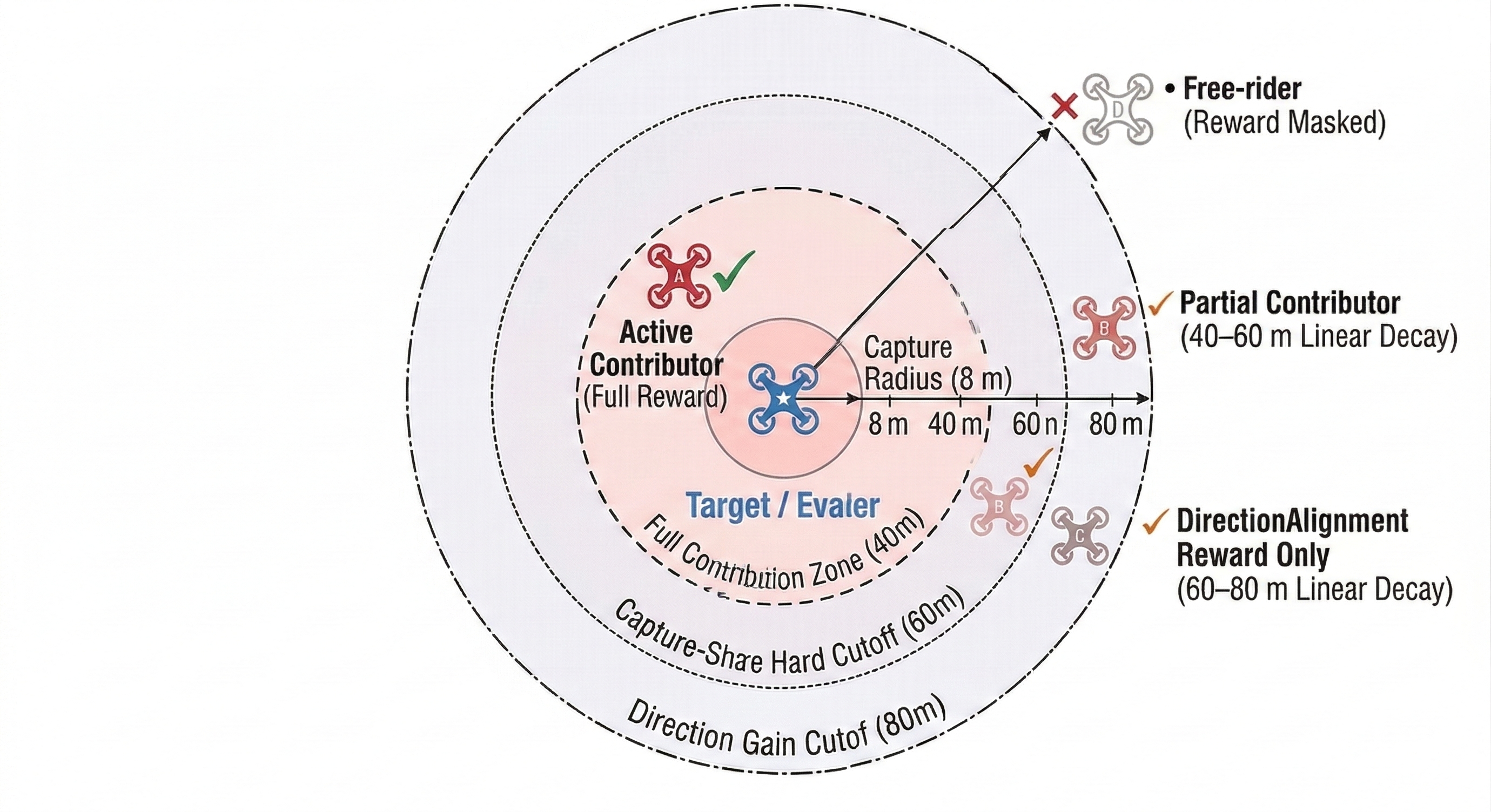}
\caption{Contribution-Gated Credit Assignment. Direction reward is fully active within 40 m, decays over 40--80 m, and vanishes beyond 80 m; capture-share is hard-gated beyond 60 m. This local credit geometry suppresses free-rider equilibria in zero-communication team pursuit.}
\label{fig:cgca}
\end{figure}

\section{Experiments and Evaluations}
\subsection{Setup and baselines}
Evaluation follows the provided Stage-5 protocol (main benchmark: 3 seeds, 500 episodes/seed; robustness suites: 200 episodes/setting/seed, with delay using 4--5 evaluation seeds). Baselines include \fullobs, \euclidean, \localnogate, \ctde, and \textsc{APF+PN}.

Comparisons are intentionally factorized. \ours, \fullobs, and \localnogate\ belong to the same inherited parameter-shared decentralized PPO family; \ours\ vs. \fullobs\ isolates information-structure effects (50-D parsimonious vs. 83-D team-coupled), and \ours\ vs. \localnogate\ isolates the role of CGCA under the same 50-D setting. \ctde\ is treated separately as a centralized-critic baseline to contrast centralized coupling against decentralized execution in this task.

\subsection{Curriculum maturation and visibility gating}
To clarify where search capability originates, we adopt a stage-wise curriculum with explicit \emph{visibility gating}. During training, target observability is conditionally released only when either the explored-space ratio exceeds a stage-specific threshold $\tau$ or a hard episode timeout is reached ($>1500$ steps in the late-stage setting). The threshold $\tau$ is progressively increased with curriculum maturation: Stage 1 has no gating, while later blind-search stages tighten from 45\% (Stage 3) to 60\% (Stage 4) and 70\% (Stage 5).

This mechanism creates a structural bottleneck: agents must first internalize 3D exploration before reliable tracking signals become available. In other words, observability is earned through coverage rather than granted a priori, which prevents early lucky sightings from dominating learning. Critically, this gate is \emph{only active during training} as an inducement for tactical emergence; it is fully removed during evaluation, where policies are tested without artificial visibility constraints.

\subsection{Training dynamics and sample efficiency}
Fig.~\ref{fig:train} shows that \ours\ reaches the high-return regime earlier and with lower oscillation amplitude than \fullobs\ and \ctde. A key reason is representation-level compression: removing 33 dimensions of delay-sensitive team-coupling features shrinks the actor input manifold and reduces nuisance covariance under delayed observations. In PPO, the policy-gradient estimator
\begin{equation}
\hat{\bm g}_t=\nabla_\theta\log\pi_\theta(a_i^t\mid\bm o_i^t)\,\hat A_i^t
\end{equation}
can be interpreted through a policy-input variance lens: fewer delay-sensitive channels reduce nuisance covariance, which is consistent with empirically more stable policy updates and improved sample efficiency in our curriculum. In contrast, \ctde\ shows less stable value-learning dynamics in this setting: the 338-D centralized critic may increase approximation burden and potentially amplify distribution-shift sensitivity once visibility gates and clutter-induced mode switches become active.

Importantly, the middle panel of Fig.~\ref{fig:train} (Clean Win Rate) shows that \ctde\ is not weak from the outset: in simpler curriculum phases (Stage 1--2), it stays above 90\% for a continuous \mbox{2M-step} interval before Stage 3. Instability emerges mainly at the Stage-3 transition, where severe 3D non-convex occlusions and visibility gating are introduced. These observations suggest that centralized critics may become fragile when high-dimensional value approximation and abrupt visibility-induced distribution shift occur simultaneously.

\textbf{Why may \ctde\ become unstable here?} The centralized critic consumes 338-D input, which may intensify approximation difficulty in highly non-convex 3D clutter. Meanwhile, visibility-gated phase transitions induce abrupt observation-distribution shifts, potentially amplifying critic-target variance and gradient jitter. The combined evidence is consistent with unstable value learning and weaker policy improvement despite centralized conditioning.

\subsection{Main benchmark and ablation study}
\begin{table}[htbp]
\caption{Main Stage-5 benchmark (mean$\pm$std over 3 seeds, 500 episodes/seed). \ours, \fullobs, and \localnogate\ are inherited parameter-shared decentralized PPO-family variants.}
\label{tab:main}
\centering
\scriptsize
\resizebox{\linewidth}{!}{
\begin{tabular}{lcccc}
\toprule
\textbf{Method} & \textbf{Success} & \textbf{Clean} & \textbf{Collision} & \textbf{Avg Steps} \\
\midrule
\ours & \textbf{0.753$\pm$0.091} & \textbf{0.752$\pm$0.091} & \textbf{0.223$\pm$0.066} & 683.6$\pm$269.4 \\
\fullobs & 0.721$\pm$0.071 & 0.721$\pm$0.072 & 0.253$\pm$0.089 & \textbf{655.4$\pm$106.7} \\
\euclidean & 0.586$\pm$0.120 & 0.583$\pm$0.121 & 0.353$\pm$0.092 & 887.9$\pm$237.7 \\
\localnogate & 0.569$\pm$0.090 & 0.568$\pm$0.089 & 0.389$\pm$0.106 & 669.1$\pm$223.2 \\
\ctde & 0.006$\pm$0.004 & 0.006$\pm$0.004 & 0.159$\pm$0.111 & 2806.8$\pm$139.1 \\
\textsc{APF+PN} & 0.125$\pm$0.007 & 0.125$\pm$0.007 & 0.875$\pm$0.007 & 124.9$\pm$2.0 \\
\bottomrule
\end{tabular}
}
\end{table}

Within the decentralized parameter-shared PPO family, \ours\ vs. \fullobs\ isolates the effect of information structure: removing explicit teammate-coupled channels improves success (0.753 vs. 0.721) and reduces collisions (0.223 vs. 0.253). This supports the hypothesis that richer coupling is not always more robust under delayed/noisy pursuit.

\textbf{Ablation insight (\ours\ vs. \localnogate).} Removing CGCA under the same 50-D zero-communication setting causes an 18.4-point success drop (0.753\,$\rightarrow$\,0.569) and a 16.6-point collision increase (0.223\,$\rightarrow$\,0.389). This isolates CGCA as a necessary mechanism rather than a cosmetic regularizer.

\subsection{Kinematic stress tests}
\begin{figure}[htbp]
\centering
\includegraphics[width=\linewidth]{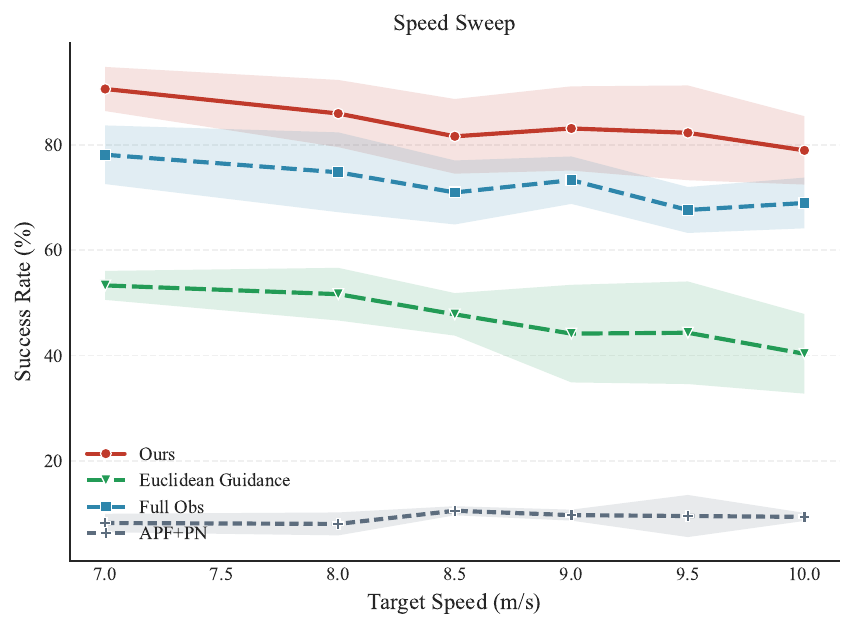}

\caption{Velocity sweep. \ours\ maintains the strongest success--collision trade-off under increasing evader speed, while \euclidean\ suffers persistent high-step inefficient pursuits.}
\label{fig:speed}
\end{figure}

Fig.~\ref{fig:speed} shows speed sweep (7.0--10.0 m/s), where \ours\ degrades gracefully from 0.907 to 0.790 success. \euclidean\ exhibits the classic kinematic trap: straight-line chasing delays feasible turn commitment in clutter, causing overshoot-recovery loops and inflated episode length.

\begin{figure}[htbp]
\centering
\includegraphics[width=\linewidth]{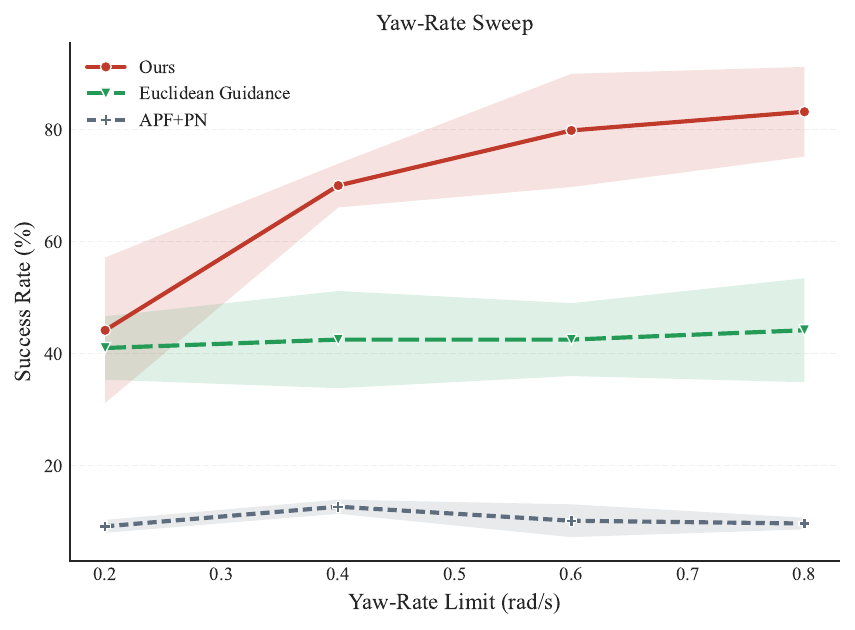}
\caption{Yaw-rate sweep. The proposed configuration preserves a robust mid-regime margin and shows graceful degradation as maneuver authority shrinks.}
\label{fig:yaw}
\end{figure}

Fig.~\ref{fig:yaw} reports yaw-cap restriction (0.8\,$\rightarrow$\,0.2 rad/s). Under severe turning limits, performance drop reflects algorithm--physics mismatch: reactive corrections arrive too late when feasible arc space collapses.

\subsection{Sim-to-real robustness}
\begin{figure}[htbp]
\centering
\includegraphics[width=\linewidth]{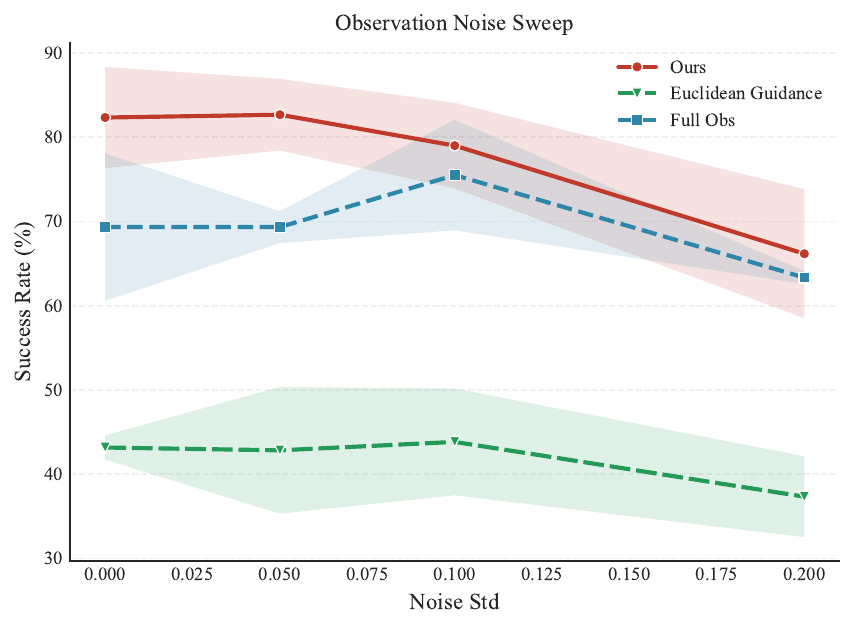}
\caption{Observation-noise sweep ($\sigma\in[0,0.20]$). \ours\ remains strongest across all tested noise levels.}
\label{fig:noise}
\end{figure}

\begin{figure}[htbp]
\centering
\includegraphics[width=\linewidth]{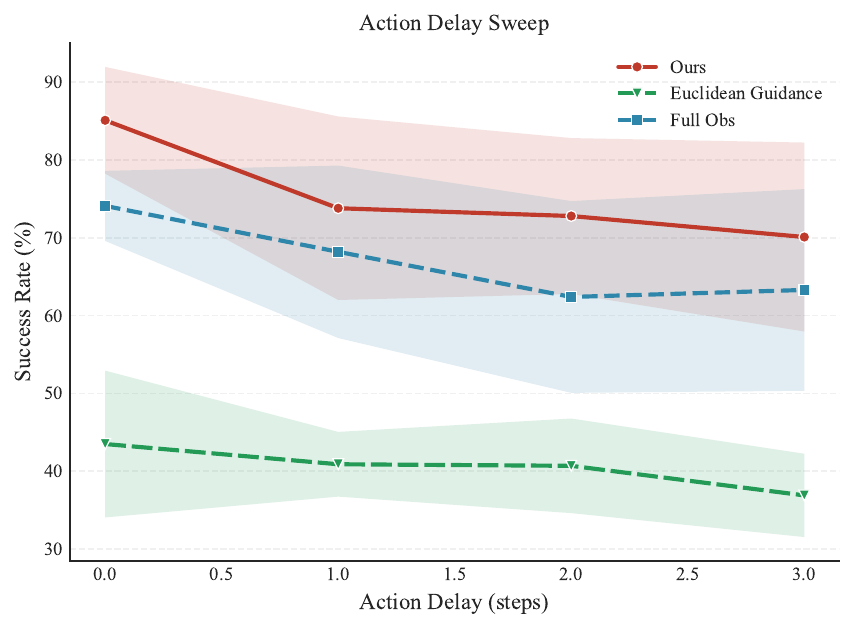}
\caption{Action-delay sweep (0--3 steps). \ours\ preserves the best absolute robustness profile under latency perturbation.}
\label{fig:delay}
\end{figure}

Noise and delay robustness are shown in Fig.~\ref{fig:noise} and Fig.~\ref{fig:delay}. Under delay, \fullobs\ degrades more sharply than \ours, which is consistent with state de-synchronization: densely coupled teammate channels may propagate stale estimates and potentially amplify control-error cascades.

\subsection{Qualitative analysis: cross-layer vertical sweep}
\begin{figure}[htbp]
\centering
\includegraphics[width=\linewidth]{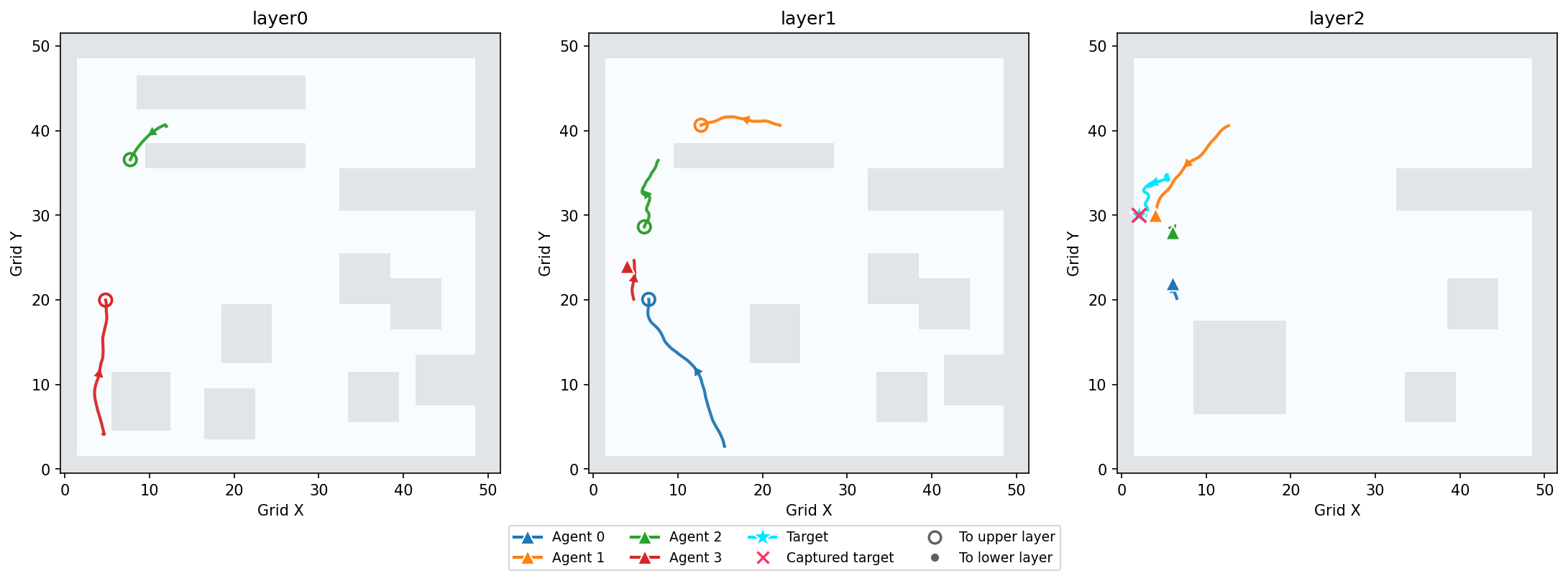}
\caption{3D Z-axis tactical slicing. The team performs cross-layer vertical sweep: upper-layer pressure constrains vertical escape while lower-layer agents close horizontal corridors, yielding a communication-free topological enclosure.}
\label{fig:qual}
\end{figure}

\begin{figure*}[t]
\centering
\includegraphics[width=0.49\linewidth]{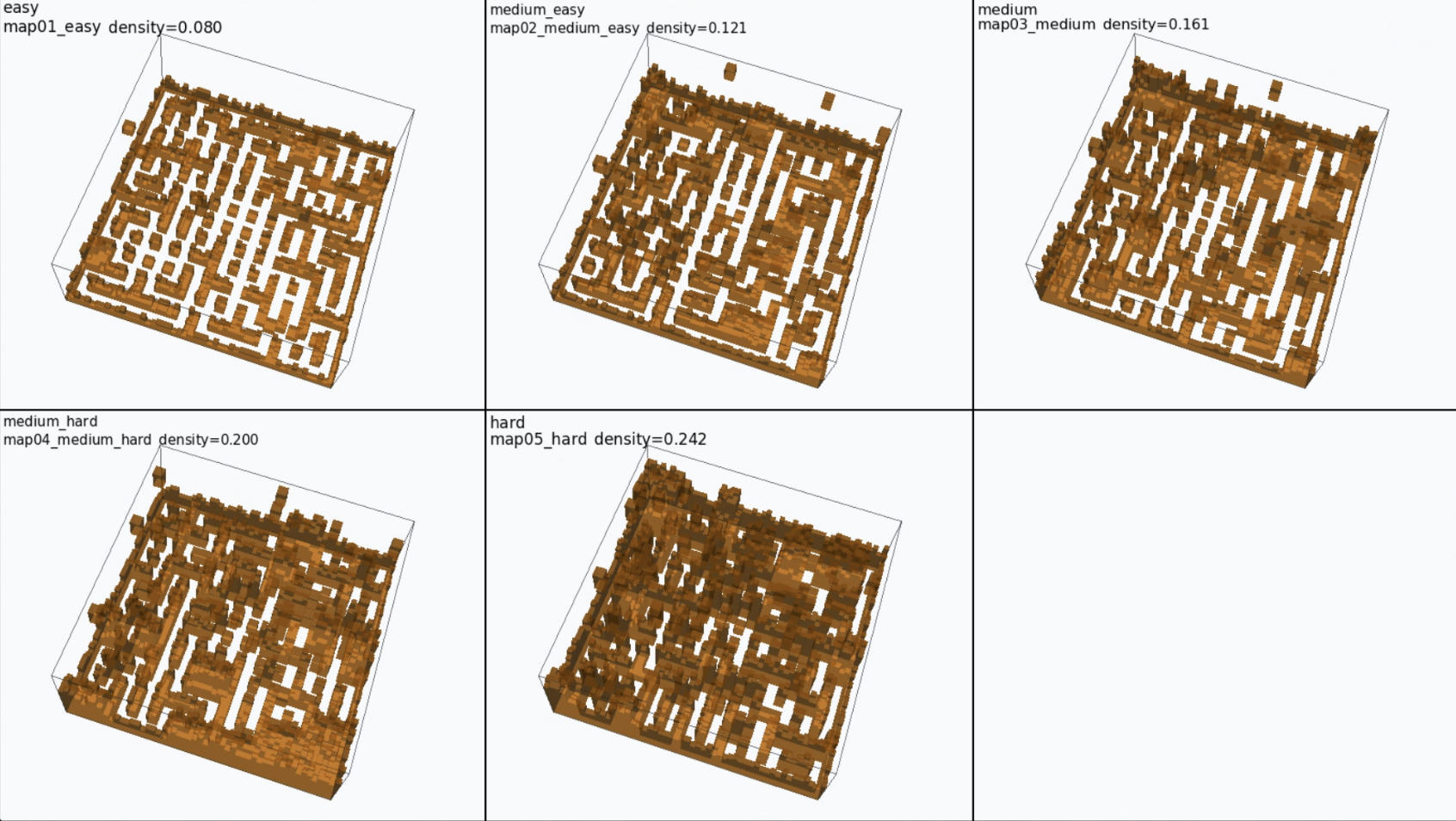}\hfill
\includegraphics[width=0.49\linewidth]{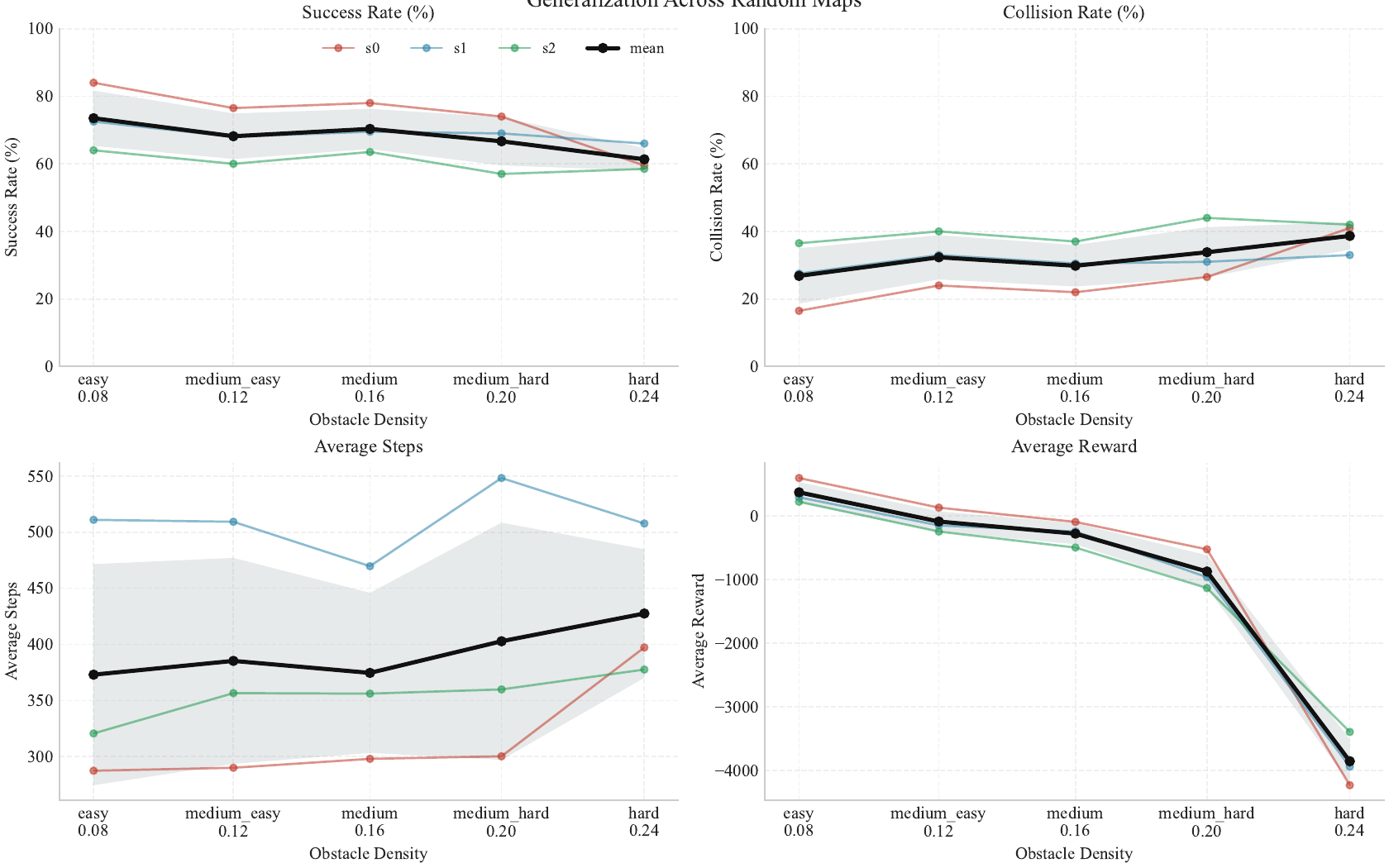}
\caption{Zero-shot transfer on five unseen Urban Canyon maps (density 0.08--0.24). \textbf{Left:} map examples from easy to hard density. \textbf{Right:} seed-wise and mean success, collision, average steps, and reward without fine-tuning. \ours\ degrades gracefully with density, remaining near 61\% success and below 40\% collision at density 0.24, consistent with structural transfer rather than fixed-map memorization.}
\label{fig:zeroshot}
\end{figure*}

Fig.~\ref{fig:qual} reveals a three-phase maneuver pattern rather than a single reactive chase. \emph{Initial Search}: guided by the shared 3D A* prior, agents fan out to frontier-consistent corridors while preserving safety spacing. \emph{Altitude Stratification}: once target likelihood concentrates, one pursuer climbs to the upper low-occupancy band and the remaining agents occupy mid/low bands, exploiting measured anisotropy (15.9\%/12.1\%/10.2\%) to reduce vertical escape options. \emph{Topological Containment}: the formation then uses obstacle morphology---especially U-shaped walls and corner pockets---as virtual teammates, tightening a cross-layer envelope until terminal capture.

This phased behavior explains why communication-free coordination remains effective: agents do not exchange explicit messages, yet shared topology priors and CGCA-induced local incentives synchronize role allocation over time. Importantly, the enclosure trajectory is geometrically consistent with the map's layered free-space structure, indicating true 3D tactical organization rather than incidental altitude jitter.

\section{Zero-Shot Generalization on Procedurally Generated Urban Canyons}
To test map over-fitting, we evaluate \ours\ zero-shot on five procedurally generated \emph{Urban Canyon} maps with obstacle densities 0.08--0.24. The generator uses connected 2.5D anisotropic street-corridor layouts (rather than random voxel noise), yielding physically meaningful maneuver corridors and controlled topology shift.

As shown in Fig.~\ref{fig:zeroshot}, performance degrades gracefully with density: at the hardest map (0.24), success remains around 61\% while collision stays below 40\%, and average steps remain in the 350--450 band. These trends are consistent with structural transfer (cross-layer containment under vertical anisotropy) rather than fixed-map memorization.

\section{Discussion and Limitations}
The empirical picture supports a robustness-oriented design principle: in communication-constrained multi-robot pursuit, less cross-agent coupling can sometimes yield more reliable coordination. In our setting, the 83-D team-coupled profile is outperformed by the 50-D parsimonious profile, and cooperation quality is maintained by locality-aware CGCA without explicit communication. This combination is consistent with the view that stale coupling channels may introduce harmful variance under delay/noise, whereas sparse local signals can remain controllable.

Two limitations remain. First, yaw-restriction sweeps currently do not include \fullobs, so low-yaw cross-method comparisons are incomplete. Second, the present noise suite is bounded at $\sigma\le0.20$; stronger perturbation regimes are needed for firmer conclusions under extreme sensing corruption.

\section{Conclusion}
Built on the inherited path-guided decentralized scaffold of \cite{pgf_mappo_2025}, this paper studies a different question from framework design: robustness-oriented information and credit structure in communication-constrained 3D pursuit--evasion. The results support two linked findings: (i) richer team-coupled observations are not always better, and representational parsimony can improve robustness; (ii) communication-free coordination can remain effective when local credit assignment is properly structured through CGCA. Across benchmark, stress, and zero-shot urban-canyon transfer evaluations, the evidence indicates a practical design principle rather than a universal rule: in communication-constrained multi-robot pursuit, reducing explicit cross-agent coupling can sometimes yield more robust coordination.


\begin{thebibliography}{99}

\bibitem{pgf_mappo_2025}
J. Ying, Z. Li, Z. Dong, G. Wu, and Y. Liao, ``Generalizable Collaborative Search-and-Capture in Cluttered Environments via Path-Guided MAPPO and Directional Frontier Allocation,'' \emph{arXiv preprint arXiv:2512.09410}, 2025.

\bibitem{ctde2017}
R. Lowe \emph{et al.}, ``Multi-agent actor-critic for mixed cooperative-competitive environments,'' in \emph{Advances in Neural Information Processing Systems}, 2017.

\bibitem{valueDecomp2018}
J. Foerster, G. Farquhar, T. Afouras, N. Nardelli, and S. Whiteson, ``Counterfactual multi-agent policy gradients,'' in \emph{AAAI Conference on Artificial Intelligence}, 2018.

\bibitem{mappo2022}
C. Yu \emph{et al.}, ``The surprising effectiveness of PPO in cooperative multi-agent games,'' in \emph{NeurIPS Datasets and Benchmarks}, 2022.

\bibitem{commnet2016}
S. Sukhbaatar, A. Szlam, and R. Fergus, ``Learning multiagent communication with backpropagation,'' in \emph{Advances in Neural Information Processing Systems}, 2016.

\bibitem{atoc2018}
J. Jiang and Z. Lu, ``Learning attentional communication for multi-agent cooperation,'' in \emph{Advances in Neural Information Processing Systems}, 2018.

\bibitem{tarmac2019}
A. Das, T. Gervet, J. Romoff, D. Bouchard, L. V. Belanger, and J. Pineau, ``TarMAC: Targeted multi-agent communication,'' in \emph{International Conference on Machine Learning}, 2019.

\bibitem{dgn2020}
Y. Jiang, Y. Guo, and S. Li, ``Graph convolutional reinforcement learning for multi-agent cooperation,'' in \emph{International Conference on Learning Representations Workshop}, 2020.

\bibitem{vdn2018}
P. Sunehag \emph{et al.}, ``Value-decomposition networks for cooperative multi-agent learning,'' in \emph{International Conference on Autonomous Agents and Multiagent Systems}, 2018.

\bibitem{qmix2018}
T. Rashid \emph{et al.}, ``QMIX: Monotonic value function factorisation for deep multi-agent reinforcement learning,'' in \emph{International Conference on Machine Learning}, 2018.

\bibitem{hart1968astar}
P. E. Hart, N. J. Nilsson, and B. Raphael, ``A formal basis for the heuristic determination of minimum cost paths,'' \emph{IEEE Transactions on Systems Science and Cybernetics}, vol. 4, no. 2, pp. 100--107, 1968.

\bibitem{khatib1986apf}
O. Khatib, ``Real-time obstacle avoidance for manipulators and mobile robots,'' \emph{The International Journal of Robotics Research}, vol. 5, no. 1, pp. 90--98, 1986.

\bibitem{zarchan2012pn}
P. Zarchan, \emph{Tactical and Strategic Missile Guidance}, 6th ed. Reston, VA, USA: AIAA, 2012.

\bibitem{sim2real2017}
F. Sadeghi and S. Levine, ``CAD$^2$RL: Real single-image flight without a single real image,'' in \emph{Robotics: Science and Systems}, 2017.

\bibitem{marlsurvey2008}
L. Busoniu, R. Babuska, and B. De Schutter, ``A comprehensive survey of multiagent reinforcement learning,'' \emph{IEEE Transactions on Systems, Man, and Cybernetics, Part C}, vol. 38, no. 2, pp. 156--172, 2008.

\end{thebibliography}
\end{document}